\documentclass[chi$_$draft]{sigchi}

\CopyrightYear{2023}

\usepackage{balance}       
\usepackage{graphics}      
\usepackage[T1]{fontenc}   
\usepackage{txfonts}
\usepackage{mathptmx}
\usepackage[pdflang={en-US},pdftex]{hyperref}
\usepackage{color}
\usepackage{booktabs}
\usepackage{textcomp}
\usepackage{dialogue}
\usepackage{multirow}
\usepackage{mathtools}
\usepackage{microtype}        
\usepackage{ccicons}          
\usepackage{todonotes}
\usepackage{blkarray, bigstrut}
\usepackage{smartdiagram}
\usesmartdiagramlibrary{additions}
\usepackage{enumitem}

\def\SYSTEM{{\em Ericson\ }}

\def\plaintitle{Ericson: An Interactive Open-Domain Conversational Search Agent}

\newcommand\commentout[1]{}

\begin{document}

\title{\plaintitle}

\numberofauthors{5}
\author{
  \alignauthor{Zihao Wang\\
    \affaddr{Mathematics \& Computer Science Department, Emory University}\\
    \affaddr{Atlanta, US}\\
    \email{zihao.wang2@emory.edu}}\\
  \alignauthor{Ali Ahmadvand\\
    \affaddr{Mathematics \& Computer Science Department, Emory University}\\
    \affaddr{Atlanta, US}\\
    \email{ali.ahmadvand@emory.edu}}\\
  \alignauthor{Jason Ingyu Choi\\
    \affaddr{Mathematics \& Computer Science Department, Emory University}\\
    \affaddr{Atlanta, US}\\
    \email{in.gyu.choi@emory.edu}}\\
  \alignauthor{Payam Karisani\\
    \affaddr{Mathematics \& Computer Science Department, Emory University}\\
    \affaddr{Atlanta, US}\\
    \email{payam.karisani@emory.edu}}\\
  \alignauthor{Eugene Agichtein\\
    \affaddr{Mathematics \& Computer Science Department, Emory University}\\
    \affaddr{Atlanta, US}\\
    \email{eugene@mathcs.emory.edu}}\\
}


\maketitle

\begin{abstract}

Open-domain conversational search (ODCS) aims to provide valuable, up-to-date information, while maintaining natural conversations to help users refine and ultimately answer information needs. However, creating an effective and robust ODCS agent is challenging. In this paper, we present a fully functional ODCS system, \SYSTEM, which includes state-of-the-art question answering and information retrieval components, as well as intent inference and dialogue management models for proactive question refinement and recommendations. Our system was stress-tested in the Amazon Alexa Prize, by engaging in live conversations with thousands of Alexa users, thus providing empirical basis for the analysis of the ODCS system in real settings. Our interaction data analysis revealed that accurate intent classification, encouraging user engagement, and careful proactive recommendations contribute most to the users' satisfaction. Our study further identifies limitations of the existing search techniques, and can serve as a building block for the next generation of ODCS agents.


\end{abstract}

\category{H.3.3.}{Information Search and Retrieval}{Open-Domain Conversational Search}



\section{Introduction}
\label{intro}
Conversational search is a crucial step towards the dream of truly conversational AI agents. Unlike in traditional Web search or automatic question answering, which typically consists of one-sided interactions of a user making a request/question and the system responding with one or more answers, a conversation could allow the user to better formulate and understand their information needs, and even help them ask the right questions by proactively recommending information relevant to the conversation context. 

More specifically, a conversational search system should include at least two capabilities: being able to retrieve information instantly upon users' requests, and proactively recommend information if the user does not know what to ask; and to maintain the state, and flow, of conversation to allow the user to naturally refine or participate in the search process. 

To guide our design of an effective conversational search system, we formulate the following research questions:
\begin{itemize}
\item RQ1: What is the system functionality needed for a satisfactory conversational search agent?
\item RQ2: How is the user satisfaction and engagement affected by the proactiveness and perfomance of individual conversational search system components?
\end{itemize}

To answer these questions, we developed and incrementally deployed our conversational agent, \SYSTEM, which allows us to observe the benefits and drawbacks of our design decisions on real user conversations. Our implementation combined state-of-the-art question-answering and web search components with conversational functionalities to enable both traditional and interactive access to relevant and interesting information. 

One benefit of having a system interact with users in the real world is that a number of limitations and directions for improvement are revealed. \SYSTEM was deployed as one of the participants in the Amazon Alexa Prize challenge\footnote{\url{https://developer.amazon.com/alexaprize}}. Alexa Prize 2017 is a challenge initiated by Amazon, which aims to build a social bot that can coherently conduct conversations with humans. While our focus was on investigating conversational search, Alexa Prize provided a venue to test our ideas in a realistic setting. Through the challenge, a large amount of interaction data, as well as explicit ratings and free-form feedback from users were collected. Using these data, we report our analysis of the key factors that affect user experience with our system, such as the accuracy of intent classification, and the ``aggressiveness'' of the system proactivity and information recommendation. Specifically, our contributions include:
\begin{itemize}
\item The development of \SYSTEM, the first fully-functional open-domain conversational search prototype, focusing on satisfying information needs while entertaining the user (Section \ref{sec:system}).
\item A case study of deploying \SYSTEM for many real users (Section \ref{sec:data}).
\item Analysis of the critical factors impacting user satisfaction and engagement in conversational search (Section \ref{sec:results}).
\end{itemize}
%
%

To our knowledge, this paper presents the first large-scale study of a fully functional open-domain conversational search system with real users, utilizing state-of-the-art search and question answering components. The system goes beyond previous published attempts at conversational search, and presents extensive analysis of the effects of the system design and component implementation choices on user satisfaction and engagement. Next, we present related work on conversational (search) agents to put our contribution in context. 

\section{Related Work}
\label{sec:related}

Conversational and information retrieval agents for various purposes have been studied for decades. Currently, the main research directions fall into two major categories. One is the application of machine learning, deep learning and reinforcement learning techniques on a specific part of an AI system. ~\cite{lemon2006isu,jafarpour2010filter} described applications of reinforcement learning, information retrieval techniques respectively, to improve the accuracy of response selection. The other one are proposals of dialogue management framework and studies on factors affecting dialogue interactions. 

~\cite{bos2003dipper} developed a dialogue system architecture integrating an information state update (ISU) mechanism. The structure helps to build the concept of a dialogue system and formalized the language for ISU. The RavenClaw system was developed by ~\cite{bohus2009ravenclaw} for task-oriented plan-based tasks. It was experimented with an online room reservation service. ~\cite{papaioannou2017combining} developed a system that combined AIML template based chatbot integrating a task-specific dialogue management system, which is trained by reinforcement learning techniques. ~\cite{ultes2017pydial} developed a multi-domain dialogue system, integrating a memory-based dialogue management mechanism and enabling applications of statistical methods in system modules. Various end-to-end data-driven models are built using deep learning methods ~\cite{yao2015attention,serban2016building}. ~\cite{li2016learning} studied the behavior of learner's proactivity, especially on asking about uncertainties. The achievements of the aforementioned studies are a variety of AI prototypes active in both research and industry, with evidence of improvement on certain part of a system by applying advanced technologies. 

Previous research achievements expanded the frontier of convesational agents research, but have important limitations which we attempted to address in this work. First, from the system perspective, our system allows the flexibility to easily extend dialogue logic and topics, compared to previous efforts that focused on limited domains or plan-based or task-oriented scnarios. Second, we investigate the effects of proactive news and topic recommendation for conversational search. Finally, we report detailed results of our system interacting with large numbers of real users without any restrictions on their behavior or interests, which significantly advance and ground our understanding of the conversational search task, and suggest promising directions for future research.

\section{System Description}
\label{sec:system}

In this section, first, we describe the architecture and implementation of the \SYSTEM system. Next, we describe the information flow within \SYSTEM, and the core components that comprise the system as an overview. Then, we describe the design and implementation of the key system components in details respectively.

\subsection{\SYSTEM Implementation as an Alexa skill}
\label{implement as alexa skill}

\begin{figure}[ht]
\centering
  \includegraphics[width=0.9\columnwidth]{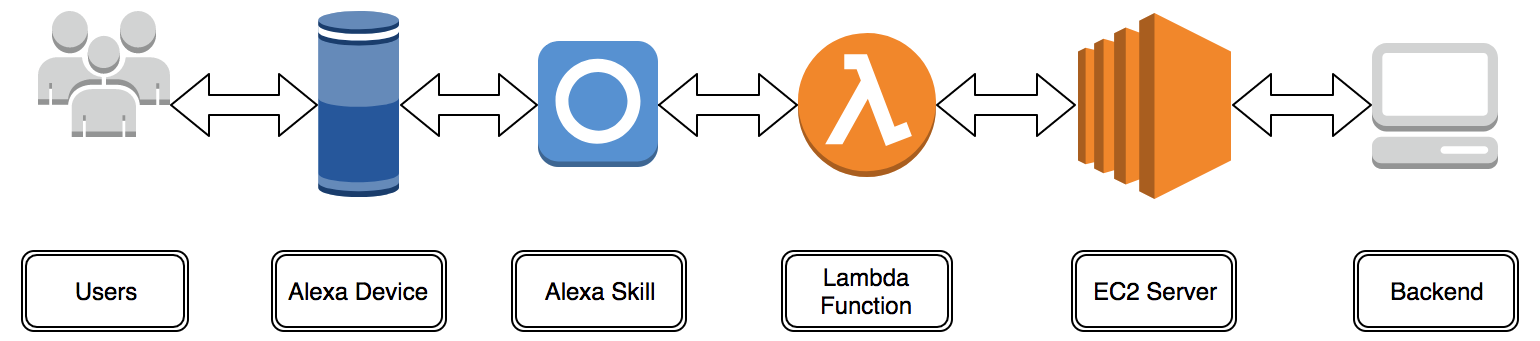}
  \caption{System deployment.}
  \label{fig:deployment}
\end{figure}

The system was deployed on Amazon EC2 servers, and connected to Alexa Skills Kit through the Lambda function ~\cite{wagner2015lambda}. \SYSTEM is activated when it is assigned to a random Alexa user and converses with the user until the conversation session ended. The connection between \SYSTEM and Alexa users is illustrated in Figure\ref{fig:deployment}.

%
%

\subsection{\SYSTEM System Overview}
\label{sec:system-overview}

\commentout{\todo{describe components IN ORDER of following the main conversation flow in overview figure}}

The \SYSTEM system is comprised of five main parts. 

\begin{itemize}
\setlength{\itemindent}{+.2in}
    \item {\em Hierarchical Intent classifier}: infers the intent of the users' utterance.
    \item {\em Dialogue State Manager (DSM)}: controls conversation flow and invokes system components.
    \item {\em Information retrieval components}: general and domain-specific question answering and search components.
    \item {\em Conversational transition component}: generates and suggests transitions to next topics to discuss.
    \item {\em Dialogue logger}: saves detailed interaction information.
\end{itemize}

\begin{figure}[h]
\centering
  \includegraphics[width=0.9\columnwidth]{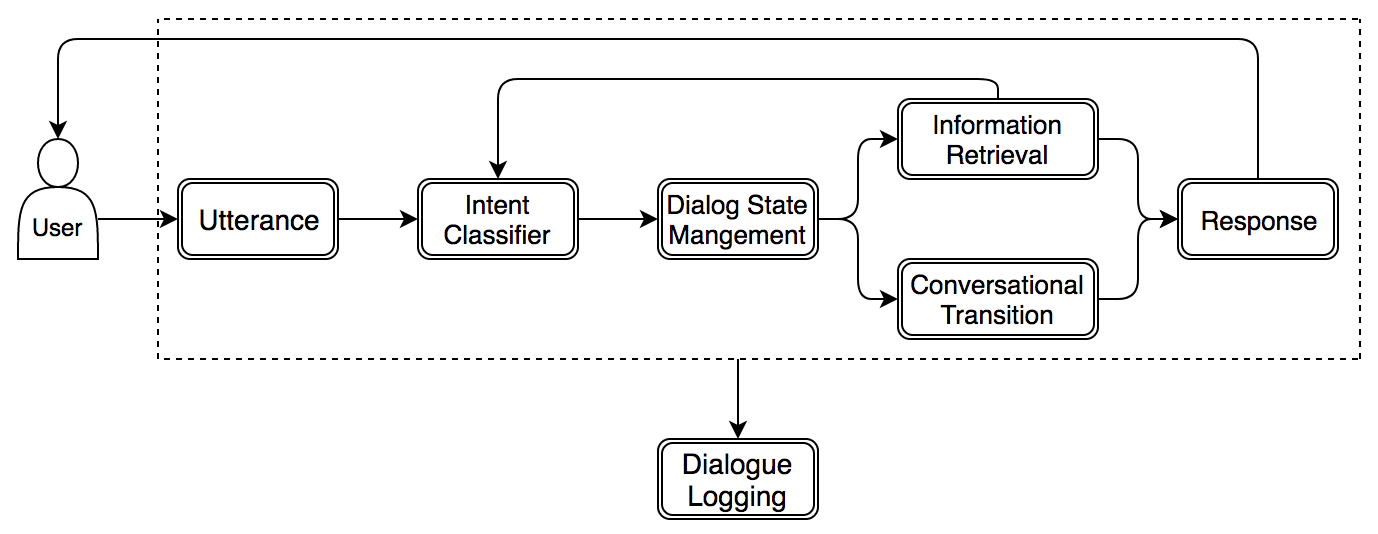}
  \caption{System overview.}
  \label{fig:architecture}
\end{figure}

An overview of the system is illustrated in Figure \ref{fig:architecture}. The intent classifier interacts with the dialogue state manager to direct the flow of the conversation. Specifically, the intent classifier tries to classify the users' intent into corresponding general or specific classes, described in detail below. The Dialogue State Manager (DSM) controls the conversational flow based on conversational logic. The DSM, based on the input from the intent classifier and the current conversational state, guides the conversation to the next topic, or continues the discussion within the current topic. When invoked by the DSM, the information retrieval components retrieve the requested information as well as potential follow-up information, as described in detail below. Similarly, based on the intent classification, the DSM may invokes a conversational transitions component provide the appropriate follow-up related topic recommendations, or offer to transition the conversation to different components or topics. Finally, dialogue logging stores all the information generated throughout the conversation. 

\commentout{\todo{Section (detailed component description) still not in same order as the overview. Subsections should follow same order as list of components above.}}

\begin{figure}[h]
\centering
  \includegraphics[width=0.9\columnwidth]{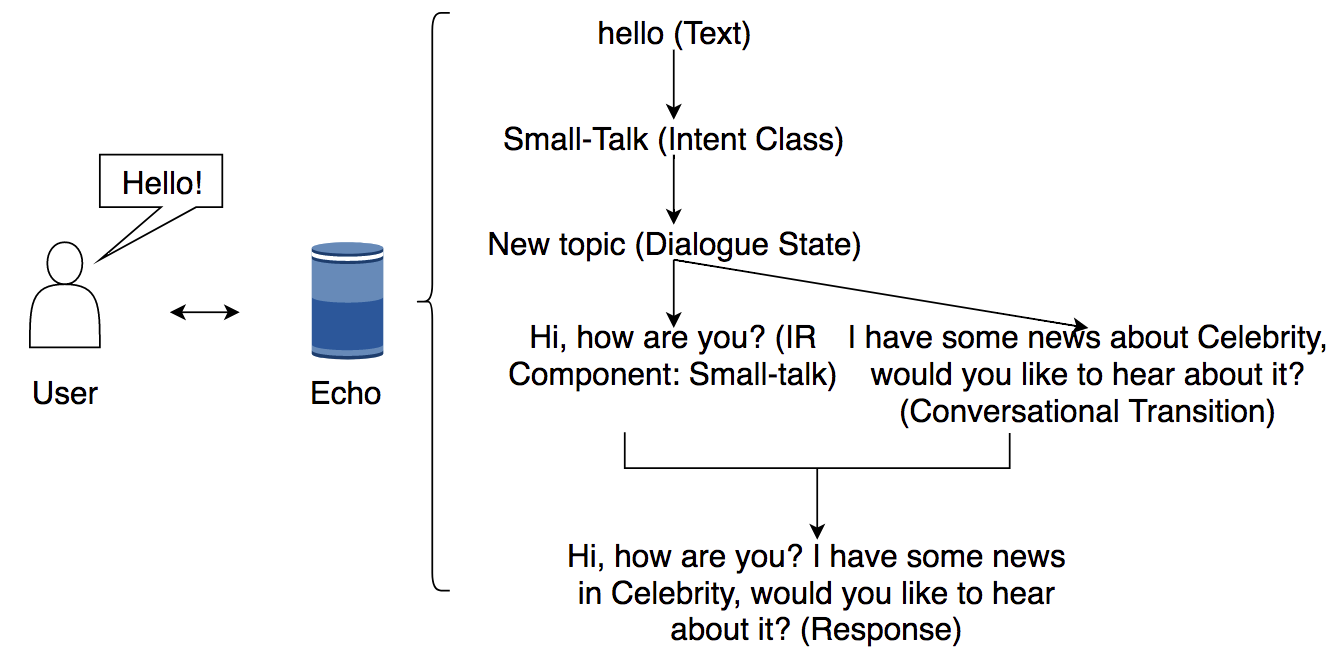}
  \caption{Conversation flow example}
  \label{fig:flow example}
\end{figure}

An example conversation is illustrated in Figure \ref{fig:flow example}. When the system is activated by a user, the intent classifier first categories the utterance, "hello", with a class label "Small-Talk". This information is checked once again by dialogue state manager based on the current dialogue state "New Topic". After this process, dialogue state manager finally determines the direction of the conversational flow and sends the utterance to the corresponding retrieval module "Small-Talk". At the end, conversational transition will incorporate the collected information, and generate a response "Hi, how are you?" from the "Small-Talk" component, and suggest a popular follow-up topic, "Celebrity", and compose a final response. As this process continues throughout the conversation, dialogue logging stores all the processed information to maintain the full convesarsation context, for use both during the conversation and for future analysis.

\commentout{\todo{Describe in text the conversation flow, using a simple example here. You can later refer to this example in subsections about each component/step in the conversation.}}
\commentout{\todo{Figure itself: Add person on left with arrow to utterance, and draw arrow from response back to person. Maybe move response down closer to person on left}}

\subsection{Hierarchical Intent Classifier}
\label{hierarchical user intent}
\begin{figure}[h]
\centering
  \includegraphics[width=0.9\columnwidth]{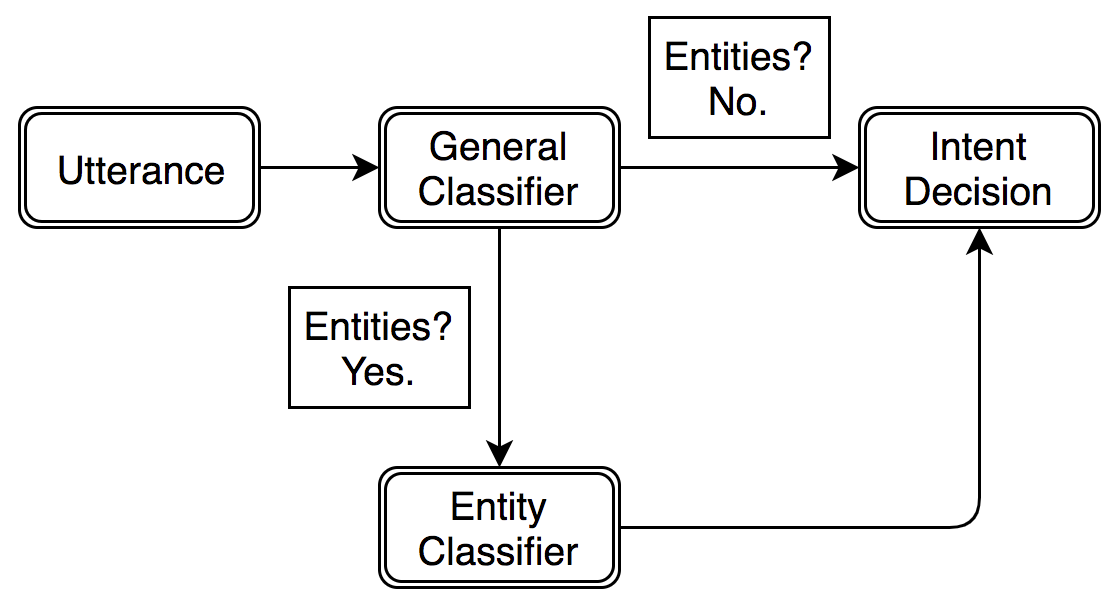}
  \caption{Hierarchical classifier.}
  \label{fig:hierarchical classifier}
\end{figure}

A hierarchical intent classification structure is formed to classify users' intent, both semantically and grammatically, to categories corresponding to sentiment (yes/no) and specific topics of interests. The structure of the classifier is shown in \ref{fig:hierarchical classifier}. 

The top level of the classifier determines which retrieval module would best fit to return the relevant information, based on the predicted topic of the utterance and the conversation context, the classifier  If the user utterance is determined to contain a specific entity, a second-level, topic-specific intent classifier is invoked, to further refine the intent category of the user. For example, utterances such as ``let's talk about Bruno Mars'' may not be classified as ``Music'', and related to Music-search component by the general classifier, but may be labeled as a ``conversation transition'' class. However, by using the entity classifier (using Google Knowledge Graph API), the entity ``Bruno Mars'' is detected and labeled as ``American singer-songwriter''. Using this information, the intent classifier can then re-classify the utterance correctly as a music related request.

Through preliminary log analysis of available conversations and popular historical Alexa requests, 14 intent classes were identified. Of those, two are sentiment classes (Positive and Negative) and 12 are the ones pertinent to topics of interest. For the classifier algorithm itself, the performances of Naive Bayes (NB), GBDT, Maximum Entropy (ME) ~\cite{nigam1999using}, and Convolutional Neural Network (CNN) ~\cite{kim2014convolutional}, were compared as candidates. Ultimatately, we chose the Gradient Boosted Decision Tree (GBDT)~\cite{friedman2001greedy} as the classifier of choice, due to both high performance and known robustness and interpretability. As features, the classifier used both surface (lexical) features, as well as seminatic features derived from the word2vec ~\cite{mikolov2013efficient} representation of the utterances. Additionally, syntactic patterns corresponding to phrases such as ``how to cook X'', and ``what is the news about Y'' are also used. The lexical, semantic, and syntactic features were combined into a single vector to represent each utterance. As the results in Section \ref{sec:main-results} show, the general top-level intent classifier performs well in most, but not all cases. 

A particularly important type of utterances is one involving entities, but is not always correctly categorized by the general classifier, as illustrated in the ``Bruno Mars'' example above. To address this problem, we defined a second-level classifier that takes advantage of the Google Knwoledge graph to incorporate domain knowledge about important entities. Specifically, the Google Knowledge Graph API is integrated to build the entity classifier. The label returned by the Knowledge Graph is matched against the available domain topics (using text-based cosine similarity). If the match of the entity type and a topic domain is above the minimum threshold, the entity is classified as the coressponding topic. Several examples of entity classifications using the Google Knowledge Graph API are shown in Table \ref{entity classification example}. As shown in Section~\ref{sec:results}, the resulting classification accuracy generally improves substantially after the additional, entity-based categorization is added.

\begin{table}[htbp]
      \centering
        
          \begin{tabular}{l|l|l}
          \toprule
          Entity & Description & Classification \\ 
          \hline
          Drake & Canadian rapper & Music \\ 
          Wonder Woman & Fictional character & Movie \\ 
          Apple & Technology company & Technology \\ 
          Donald Trump & 45th US President & News \\ 
         \bottomrule
         \end{tabular}
       \caption{Knowledge Graph API and Entity Classification Example}
       \label{entity classification example}
\end{table}


\subsection{Dialogue State Manager(DSM)}

\begin{figure}[h]
\centering
  \includegraphics[width=0.9\columnwidth]{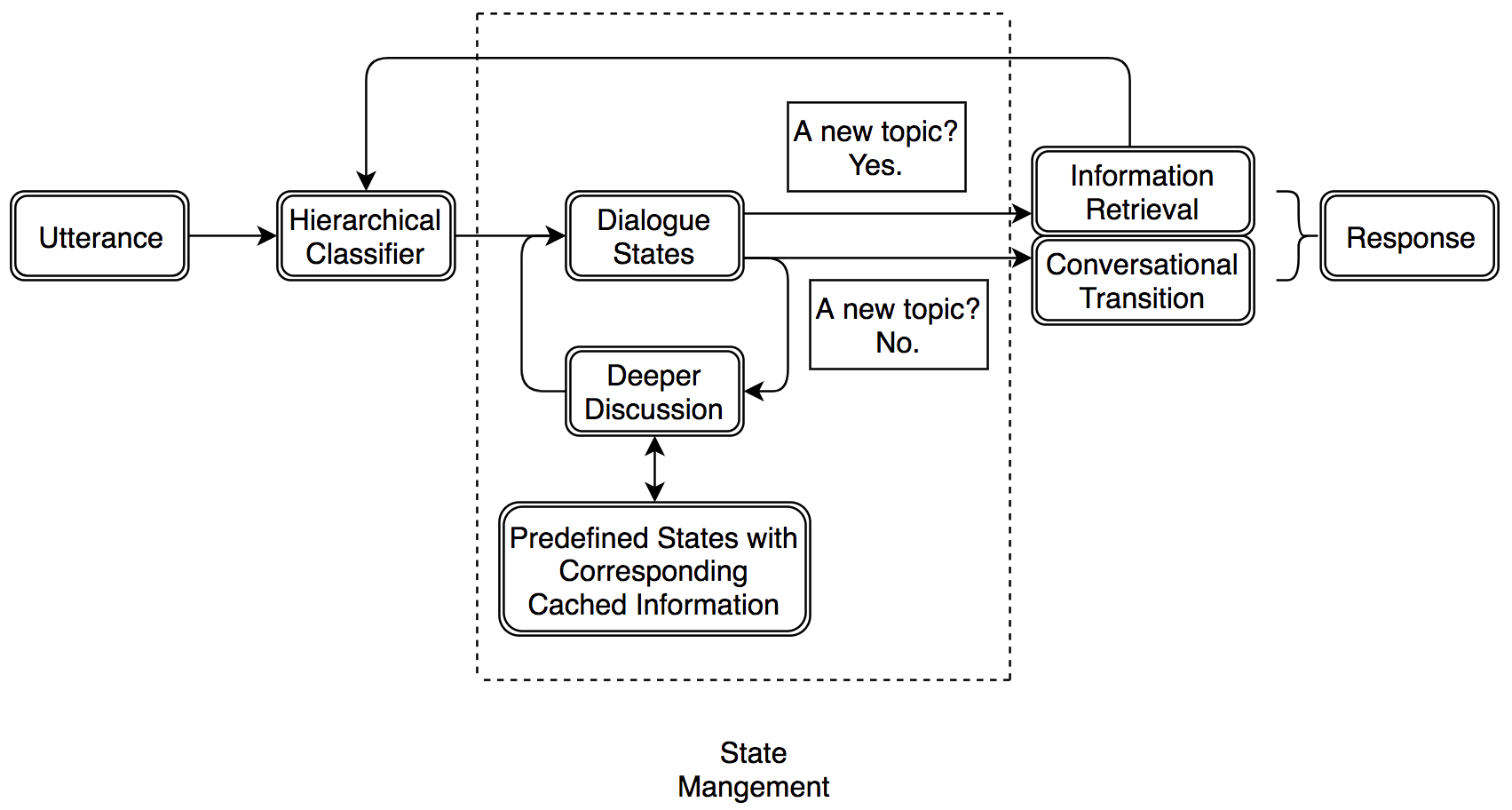}
  \caption{Dialogue State Manager.}
  \label{fig:dialogue state management}
\end{figure}

The Dialogue State Manager (DSM) plays a critical role in determining the system response. As input, the DSM uses both the intent classifier prediction as well as the conversation context, to better interpret the intent, and can interact with the user to confirm the intent if unclear. The information flow of the dialogue state manager is illustrated in Figure \ref{fig:dialogue state management}.

There are 14 top-level predefined dialogue states, such as "Greetings", "Info-request", "Movies", "Music" and so on. Within each of the dialogue state, there may be other sub-states within the top level state. As each utterance is categorized by the intent classifier to the appropriate state, the DSM maintans a stack of recent conversation states to enable back-tracking seemlessly.

When users start on a new topic, the dialogue state is only determined by the intent classifier and the state is updated in accordance with the classification result. Once users intend to discuss more details about a topic, pre-defined states predominate and the intent classifier stands by. Details of the discussion will be handled based on specific scenarios in each state, such as available information in a moview reviw site such as ``Rotten Tomatoes''. The general classifier is utilized if a user's intent is out of scope of pre-defined scnarios, and is passed to one of the (general) information retrieval components, described next.

As an example, consider the following (common) scenario. When a user starts to discuss movies, a state variable is set as "Movies", and a context variable stores all the utterances and responses from that user and the system. A user can continue talking about recent popular movies, and information would be retrieved from the cache variable, and at the same time, any pronouns can be translated by co-reference based on information in context variables. At any point, the user can switch to another topic by saying an utterance about something else.

\subsection{Information Retrieval Components}

Providing relevant and up-to-date information is a top priority of our ODCS system. Our system \SYSTEM integrates both real-time, and static sources in the information retrieval process. There are 12 information retrieval components in total, as described below. 


\commentout{\begin{table}[htbp]
      \centering
        \caption{Information Retrieval Components}
        \label{IR components}
          \begin{tabular}{l|l}
           \toprule
          Small-task & News \\ 
          Wiki-info & Weather \\ 
          Jokes & LiveQA \\ 
          Movies & Music \\ 
          Opinions & Food \\
          Transition & Unrecognized \\
          \bottomrule
         \end{tabular}
\end{table} }

\begin{itemize}
\item Small-Talk. The Small-Talk module retrieves information from a static source. It is based on modifications of AIML language templates from Alicebot ~\cite{marietto2013artificial}. This module supports dialogues in greetings, some factoid questions and simple transitional dialogues.

\item News. The News module is a real-time information retrieval module. It retrieves up-to-date news from variety of mainstream media sources such as Washington Post, CNN and ESPN.

\item Wiki-info. The Wiki-info module is a real-time information retrieval module. It retrieves summarized information from Wikipedia by the Wikipedia API. It is real-time since Wikipedia is a source that always keeps information updated. 

\item Weather. The Weather module is real-time information retrieval module. It provides temperature and weather information for a particular city from Yahoo! Weather API.

\item Jokes. The Jokes module is static. It selects random jokes from a database. This component is designed to cheer people up based on users' negative responses to greetings and entertain users upon requests.

\item LiveQA. The LiveQA module is both static and real-time. The static database is the Yahoo! Webscope Comprehensive Questions and Answers up to 2006. By using similarity matrix,the most similar question is selected and the corresponding answer is returned. Answers for How-To questions are retrieved from the Wikihow source, which is constantly updated.

\item Movies. The Movies module is both static and real-time. It retrieves relevant movie information such as actors, plots and reviews from IMDB database for relatively old movies. Recent box-office movie information is retrieved from Rotten Tomatoes.

\item Music. The Music module is real-time. It retrieves relevant music information such as ranking charts, artists information and music news from Billboard. 

\item Opinions. The Opinions module is currently unavailable due to issues of providing personal opinions on sensitive fields such as politics or religion.

\item Food. The Food module is real-time. It retrieves recipe information, such as directions, ingredients and nutrition, from Epicurious.com.

\item Transition. The Transition module is special. This component is not designed as an information retrieval component, but rather a function that directs conversation to a desired component or mechanism, when an utterance is entity related, or a recommendation is proposed. This component is crucial to drive the conversation fluently and consistently.

\item Unrecognized. The Unrecognized module is a state that a given utterance cannot be classified due to speech recognition or severe grammatical errors. A repetition request will be sent to the user.

\end{itemize}

\subsection{Conversational Transition}

Conversational transition aims to smooth the conversation flow, as well as progress through the conversation to new topics. It is composed of two functions, ``hot topic recommendation'' and ``reminders'' of available topics. 

Hot topic recommendation is based on two mechanisms. First, when users mention entities, which are identified by the entity classifier, the system will try to recommend a related topic. For example, after a user finishes talking about Bruno Mars, details about other celebrity information will be suggested if the user agrees to this topic suggestion. A list of available topics to recommend is reported in Table \ref{recommendation topics}.

\begin{table}[htbp]
      \centering
          \begin{tabular}{l|l}
           \toprule
          Basketball & Hockey \\ 
          Soccer & Animal \\ 
          Football & Science \\ 
          Baseball & Space \\ 
          Health & Technology \\
          Celebrity & Travel \\
          \bottomrule
         \end{tabular}
      \caption{Recommendation Topics}
      \label{recommendation topics}
\end{table} 

A topics reminder is called when a user continuously refuses the recommended topics. System will repeat the available topics and let users choose their desired topic. The "Suggestion" state is designed to indicate this transition state. In this state, recommendation mechanism works together with the intent classifier. The system is expecting positive or negative responses to the recommendations. If the classifier categorizes the utterance to another component rather than positive or negative, conversation transitions to the (other) the identified component. 

\subsection{Dialogue Logger}
The dialogue logger stores all the information generated through the conversation. It includes three functions, action, context, and utility.
\commentout{\begin{figure}[h]
	\centering
    \resizebox{.4\linewidth}{!}
    {%
    	\smartdiagram[constellation diagram]{Dialogue \\ Logging, Action, Context, Utility}
    }
    \caption{Dialogue logging.}
    \label{fig:dialogue logging}
\end{figure}}

\begin{itemize}

\item Action. Action is the function that tracks and stores responses from designated information retrieval components, and eventually returns the best possible answer to users.

\item Context. Context stores all dialogue state related information in variables, required for tracking the progress of conversations. There are three types of variables, state variables, cache variables, and context variables.

State variables are predefined variables to indicate different states of a conversation. Based on different information retrieval components and their capabilities of extending discussions, different states are correspondingly planned. State variables are checked and updated per turn of conversation.
Cache variables store detailed and categorized information of a topic and can be retrieved later for deeper topic discussion. 
Context variables store all the utterances and responses in sequence. Context variables can also be retrieved, mainly for the use of context understanding, such as co-reference resolution function to resolve ambiguous pronouns.


\end{itemize}

\section{Experimental Setup}
\label{sec:data}
In this section, details about performance evaluation metrics, users, experimental period, and data collection are provided.

The user-\SYSTEM interactions were conducted during the Alexa Prize 2017, with the system as a deployed Alexa skill . Alexa Prize 2017 is a competition, initiated by Amazon to university students, with the objective of developing a socialbot to conduct natural and engaging conversations with humans, in popular topics, such as entertainment, fashion, politics, sports, and technology, while providing direct feedback from users to participants to further improve the socialbot\footnote{\url{https://developer.amazon.com/alexaprize/faq}}. All socialbots in the competition are promoted to Alexa users, who are encouraged to interact with these socialbots, and at the same time, provide feedback. All socialbots and users are anonymized during the contest, and socialbots are assigned to users randomly for interactions, to ensure the objectivity of ratings. In addition, socialbots are only allowed for topic-wise conversations, and prohibited from collecting personal information to avoid confidentiality violations. 

\commentout{\todo{Again, Again! add general description of Alexa prize -- your bot was deployed as part of it! reference Alexa prize URL for more details.}}

\commentout{\todo{Rewrite completely to be self-contained! DO NOT REFER TO ALEXA PRIZE MILESTONES at all. this is a research paper that should stand independent of Alexa prize, do not assume reader is familiar with Alexa prize!}}


\subsection{Performance Evaluation Metrics}
\label{sec:metrics}

\subsubsection{Explicit User Ratings}
The primary performance metrics are the explicit user ratings to the system. Each user was asked to give a numeric rating score, ranging from 1 to 5 (with 5 meaning ``excellent'' and 1 meaning ``poor''), based on their satisfaction of the conversation, after they finished their interactions with Ericson. The rating scores were collected by the Alexa organizers.
\commentout{\todo{Describe how the explicit ratings were collected, when, scale, questions asked, and sub-criteria.}}

\subsubsection{Number of Conversation Turns}
Each conversation contains one or more turns. A conversation turn is defined as one utterance from a user and one response from the system. The number of conversation turns was considered a secondary factor in the Alexa Prize competition, but can be a useful tool to analyze the amount of effort, interest, and engagement a user is willing to spend with the system overall, and specific components. 

\subsubsection{User Engagement}
Engagement is defined as the ability to conduct coherent conversations on a specific topic. In Ericson, several components are well developed focusing on deeper discussions about news, foods, and news. Therefore, it is measured by the number of conversation turns conducted within the specific component continuously in a conversation. 
\commentout{\todo{I did not see any engagement statistics and analysis - what happened to number of turns, length of conversation, and correlation of these to explicit ratings? In particular, can you report median number of turns within each topical component, e.g., in News, in Movies, etc?}}

\subsubsection{System Proactivity}
Proactivity is defined as the ability to drive the conversation forward by proposing or reminding of related topics to the current conversation. It is measured from two aspects, the overall ratings and the number of conversation turns with respect to the overall system.

\subsection{Users}

The users came from two populations: the regular group of Alexa users in the States, and staff members of the Amazon Alexa team. No demographic information about users are provided due to privacy protection.

\subsection{Experimental Period}

The reported data were collected from June 15th,  when the system was relatively stable, to August 30th, 
\commentout{\todo{what is "semifinals period"? See above.}}
\SYSTEM was operating live to interact with real users. 46 days of interaction data were collected, as well as user ratings and free-form feedback provided by Amazon. 

\subsection{Data Collection}
\label{sec:data-details}
The results in this paper are based on over 22,000 sessions, 230,000 turns of conversations. Users were encouraged to interact with the system and provide ratings and free-form feedback once the session ended. Due to confidentiality agreements, all the conversation examples in this paper are recreated by our researchers talking to \SYSTEM in a similar context to real dialogues. An example of the conversation is shown below.
\commentout{\todo{Did you rephrase according to Amazon response to your paper?}}
\commentout{\todo{In the end of discussion section, include new subsection on QUALITATIVE analysis, i.e., some common compliments/good comments and common criticisms/bad comments that could be informative for improving the next generation of systems}}


\commentout{

\begin{table}[htbp]
  \caption{Conversation ID, time, and score}
  \label{conv-id}
  \centering
  \begin{tabular}{l|l|l}
    \toprule
    Conversation ID & Time stamp  & Score \\
    98b7b715...427e7bf3 & 2017-07-22...07:00  & 3.0 \\
    \bottomrule
  \end{tabular}
\end{table}
}

\line(452,0){240}
    \begin{dialogue}
        \speak{User} Let's talk about Wonder Woman.
        \speak{\SYSTEM} Here are some other recent Action movies like the wonder woman. The Hitman's Bodyguard, The Adventurers, Dave Made a Maze, Fairy Tail, and Kidnap. I have more details about the movie Wonder Woman. Would you like to know about that?
        \speak{User} Yes please.
        \speak{\SYSTEM} I have the following information about Wonder Woman on plot, star, producer, rating, director and genre. Which information are you interested in?
        \speak{User} Tell me about the rating.
        \speak{\SYSTEM} Here is the rating information, 8.4. I still have plot, star, producer, director and genre. What information would you like?
        \speak{User} Director information.
        \speak{\SYSTEM} Here is the director information, Patty Jenkins. I still have plot, writer, producer and genre information. What information would you like?
        \speak{User} None of them.
        \speak{\SYSTEM} OK, I still have information about most recent movies from rotten tomatoes. Would you like to hear about that?
        \speak{User} No thanks, tell me some recent jazz music instead.
    \end{dialogue}
\line(464,0){240}

\section{Results and Discussion}
\label{sec:results}

\commentout{\todo{Include section overview first paragraph. Postpone justification and discussion to the actual "Discussion" subsection.}}

In this section, we initially present the results of the user ratings to the overall system, then, analyze the effects of the intent classifier, user engagement, and system proactivity based on the interaction data and user feedback to highlight the crucial factors on user-\SYSTEM interaction. Insights of the analysis will shed light on the directions of improvement. We finally, conclude the section with qualitative analysis of open user feedback (comments) received from a subset of users.

\subsection{Quantitative Results}
\label{sec:main-results}

\begin{figure*}[ht]
  \centering
  \includegraphics[width=180mm]{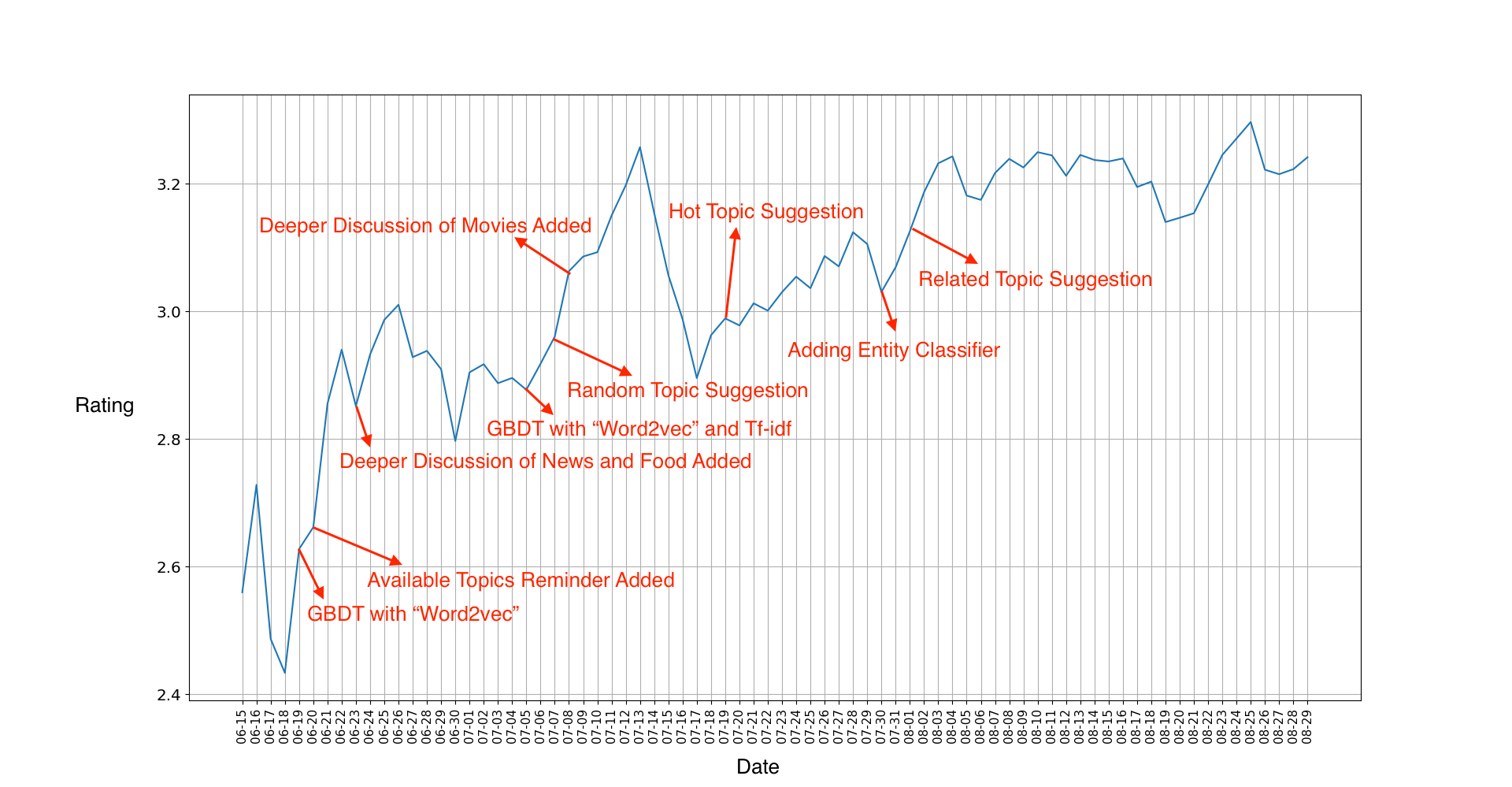}
  \caption{Overall ratings with development milestones.}
  \label{fig:overall ratings milestones}
\end{figure*}

The overall rating of \SYSTEM was improved from about 2.4 to 3.2 during the experimental period. The three factors that contributed critically to the overall improvement are intent classification, proactivity, and engagement. Daily averaged performance ratings over the period from June 15th to August 29th are shown in Figure \ref{fig:overall ratings milestones}, with indications to the milestones when major modifications were made. Table \ref{development timeline} summarizes a chronicle of the modifications in \SYSTEM.

\commentout{\todo{Figure 6 is *unreadable* -- too small! Make bigger size, at least full column! Cut other text or figures if necessary.}}



\begin{table}[htbp]
\centering
\begin{tabular}{r l}
  \toprule
  2017-6-19 & GBDT classifier + Word2Vec \\
  2017-6-20 & Available Topics Reminder \\
  2017-6-23 & Engagement within news and food module \\
  2017-7-05 & TF-IDF feature is added to GBDT \\
  2017-7-07 & Random topic suggestion \\
  2017-7-08 & Engagement within movie module \\
  2017-7-19 & Hot topic suggestion \\
  2017-7-30 & Entity Classifier \\
  2017-8-01 & Smart topic suggestion and proactivity\\
  \bottomrule
\end{tabular}
\caption{Development Timeline}
\label{development timeline}
\end{table}

\subsubsection{Hierarchical Intent Classification}

Our analysis shows that the hierarchical intent classification is crucial to the performance of the system. The classifier includes two levels of classification, the general intent classifier, and the entity classifier. During the development, as discussed in Section \ref{hierarchical user intent}, different versions of the intent classifier were integrated into the system. Figure \ref{fig:overall ratings milestones} shows that there is a noticeable improvement in the average rating, each time the modification was made. The first level of the classifier is a general purpose classifier. To evaluate this classifier, we created a dataset consisting of 4593 user utterances and question syntactic structures (e.g. "how to cook X"). Table \ref{intent_classifier_accuracy} summarizes the results (two-fold cross validation was carried out to do the experiments).

\begin{table}[htbp]
	\centering
	\fontsize{6pt}{6pt}
    \selectfont
	\begin{tabular}{@{}l|ll|ll|ll|ll@{}}
  		\toprule
  		\multirow{2}{*}{\textbf{Class Name}} & \multicolumn{2}{c} {\textbf{NB}} & \multicolumn{2}{c}{\textbf{ME}} & \multicolumn{2}{c}{\textbf{CNN}} & \multicolumn{2}{c}{\textbf{GBDT}}\\ \cmidrule(lr){2-3} \cmidrule(lr){4-5} \cmidrule(lr){6-7} \cmidrule(lr){8-9}
        
  		& \textbf{ACC} & \textbf{F1} & \textbf{ACC} & \textbf{F1} & \textbf{ACC} & \textbf{F1} & \textbf{ACC} & \textbf{F1}\\
  		\midrule
 		\textbf{Positive(S)}  & 0.281 & 0.350 & \textbf{0.861} & 0.763 & 0.724 & 0.718 & 0.825 & \textcolor{red}{\textbf{0.768}}\\

  		\textbf{Negative(S)}  & 0.584 & 0.449 & 0.759 & 0.813 & 0.832 & \textcolor{red}{\textbf{0.864}} & \textbf{0.837} & 0.846\\
\hline
  		\textbf{Small Talk (IR)} & 0.438 & 0.395 & 0.559 & 0.589 & 0.585 & 0.596 & \textbf{0.611} & \textcolor{red}{\textbf{0.631}}\\
  
  		\textbf{News (IR)} & 0.947 & 0.768 & \textbf{0.954} & 0.839 & 0. 85 & 0.849 & 0.928 & \textcolor{red}{\textbf{0.887}}\\

  		\textbf{Wiki (IR)} &  0.607 & 0.584 & 0.716 & 0.762 & \textbf{0.859} & \textcolor{red}{\textbf{0.873}} & 0.750 & 0.782\\
 
  		\textbf{Weather (IR)} & \textbf{0.970} & 0.899 & 0.938 & 0.903 & 0.875 & 0.838 & 0932 & \textcolor{red}{\textbf{0.905}}\\
  
  		\textbf{Joke (IR)} &  0.948 & 0.835 & \textbf{0.977} & \textcolor{red}{\textbf{0.931}} & 0.944 & 0.902 & 0.954 & 0.909\\
  
  		\textbf{LiveQA (IR)} &  0.339 & 0.467 & 0.685 & 0.716 & \textbf{0.735} & 0.759 & 0.713 & \textcolor{red}{\textbf{0.760}}\\
  
  		\textbf{Movies (IR)} &  0.893 & 0.661 & 0.845 & 0.889 & 0.865 & 0.893 & \textbf{0.894} & \textcolor{red}{\textbf{0.898}}\\
  
  \textbf{Music (IR)} & 0.895 & 0.676 & 0.890	& \textcolor{red}{\textbf{0.920}} & 0.832	& 0.864	& \textbf{0.898}	& 0.905\\
  
  		\textbf{Opinion (IR)} & 0.267 & 0.377 & 0.830 & 0.770 & 0.846 & 0.776 & \textbf{0.866} & \textcolor{red}{\textbf{0.788}}\\
  
  		\textbf{Food (IR)} & \textbf{0.981} & 0.633 & 0.931 & 0.867 & 0.876 & 0.827 & 0.942 & \textcolor{red}{\textbf{0.868}}\\
\hline  
  		\textbf{Transition(T)} & 0.811 & 0.614 & 0.740 & 0.769 & \textbf{0.834} & \textcolor{red}{\textbf{0.806}} & 0.788 & 0.802\\
  
  		\textbf{Unrecognized(T)} & 0.162 & 0.250 & \textbf{0.485} & 0.423 & 0.469 & \textcolor{red}{\textbf{0.487}} & 0.302 & 0.473\\
  		\bottomrule
	\end{tabular}
	\caption{Comparison of the general classifiers. NB stands for Naive Bayes classifier, ME stands for Maximum Entropy classifier, CNN stands for Convolutional Neural Network classifier, and GBDT stands for the Gradient Boosted Decision Tree classifier (with word embeddings and TF-IDF features)}.
    \label{intent_classifier_accuracy}
\end{table}


In Table \ref{intent_classifier_accuracy}, the best accuracy and F1-score for each class are shown in bold and red colors respectively. There are three groups of intent classes, Sentiment (S), Information Retrieval (IR), and Transitions (T). The results show that GBDT outperforms the other classifiers in S and IR groups, however, CNN is slightly better in T group.

To evaluate the effectiveness of the entity classifier, we uniformly sub-sampled conversations in different lengths to ensure the samples are not biased. The samples are manually annotated and used to calculate the F1 score and accuracy for each class. A more detailed comparison result is summarized in Table \ref{before and after entity classifier}.

\begin{table}[htbp]
\centering
\small
\begin{tabular}{@{}l|ll|ll@{}}

  \toprule
  \multirow{2}{*}{\textbf{Class Name}} & \multicolumn{2}{c}{\textbf{Before}} & \multicolumn{2}{c}{\textbf{After}}\\ 
  \cmidrule(lr){2-3} \cmidrule(lr){4-5}
  & \textbf{ACC} & \textbf{F1} & \textbf{ACC} & \textbf{F1}\\
  \midrule
  \textbf{Positive (S)} & 0.883 & 0.931 & 0.973 & 0.985\\
  
  \textbf{Negative (S)} & 0.899 & 0.923 & 0.972 & 0.981\\
  \hline
  \textbf{Small Talk (IR)} & 0.888 & 0.837 & 0.888 & 0.842\\
  
  \textbf{News (IR)} & 0.865 & 0.915 & 0.971 & 0.963\\
  
  \textbf{Wiki (IR)} & 0.577 & 0.695 & 0.722 & 0.812\\
  
  \textbf{Weather (IR)} & 0.933 & 0.965 & 0.960 & 0.979\\
  
  \textbf{Joke (IR)} & 0.948 & 0.973 & 0.962 & 0.980\\
  
  \textbf{LiveQA (IR)} & 0.825 & 0.812 & 0.838 & 0.818\\
  
  \textbf{Movies (IR)} & 0.900 & 0.920 & 0.956 & 0.954\\
  
  \textbf{Music (IR)} & 0.904 & 0.924 & 0.968 & 0.969\\
  
  \textbf{Opinion (IR)} & 0.750 & 0.919 & 0.770 & 0.977\\
  
  \textbf{Food (IR)} & 0.948 & 0.950 & 0.950 & 0.982\\
  \hline
  \textbf{Transition (T)} & 0.876 & 0.871 & 0.875 & 0.975\\
  
  \textbf{Unrecognized (T)} & 0.896 & 0.865 & 0.904 & 0.966\\
  \bottomrule
\end{tabular}
\caption{Results before and after adding the entity classifier}
\label{before and after entity classifier}
\end{table}

The results show that both F1 and accuracy score significantly improved after adding the entity classifier as a second-level classification system. For all S and IR groups, the hierarchical classification system outperforms the one-level classification system.

\commentout{\todo{Table 5: Organize into sub-headings of table, by intent type (domain vs. sentiment vs. conversation. Same comments as for table 1: what is important here? what are we supposed to learn from all these numbers? Which numbers are important? }}

\subsubsection{Engagement}

\commentout{\todo{Very interesting results. General comment: for each figure reported, also discuss whether change was statistically significant. Assume normal distribution around means if you don't have time to calculate proper statistics.}}

Engagement is another important factor that distinguishes how meaningful a conversation between a human and a computer is. Throughout live conversations, because 'Movies', 'Music' and 'Food' retrieval components were the most popular among all information retrieval components, we decided to expand these components and study the relationship of engagement to users' satisfaction. Here is an example conversation that shows how engagement works within the "Movies" component.

\line(559,0){240}
    \begin{dialogue}
        \speak{User} Let's talk about Wonder Woman.
        \speak{Ericson} Here are some other recent Action movies like the wonder woman. The Hitman's Bodyguard, The Adventurers, Dave Made a Maze, Fairy Tail, and Kidnap. I have more details about the movie Wonder Woman. Would you like to know about that?
        \speak{User} Yes please.
        \speak{Ericson} I have the following information about Wonder Woman on plot, star, producer, rating, director and genre. Which information are you interested in?
        \speak{User} Tell me about the rating.
        \speak{Ericson} Here is the rating information, 8.4. I still have plot, star, producer, director and genre. What information would you like?
        \speak{User} Director information.
        \speak{Ericson} Here is the director information, Patty Jenkins. I still have plot, writer, producer and genre information. What information would you like?
        \speak{User} None of them.
        \speak{Ericson} OK, I still have information about most recent movies from Rotten Tomatoes. Would you like to hear about that?
        \speak{User} No thanks, tell me about nice recent jazz songs.
    \end{dialogue}
\line(572,0){240}



\begin{figure}[h]
\centering
  \includegraphics[width=1.0\columnwidth]{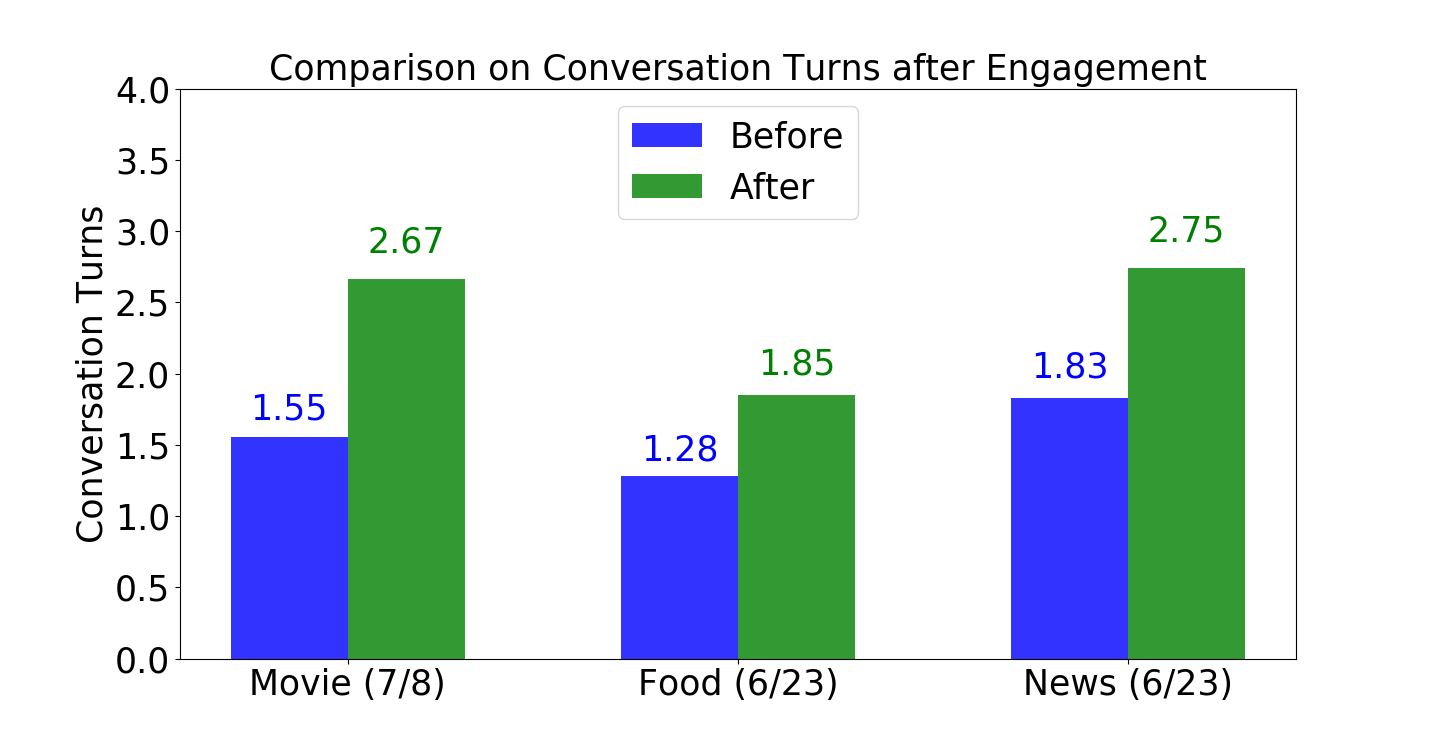}
  \caption{The average number of conversation turns within a component.}
  \label{fig:Engagement}
\end{figure}

This example shows how user can engage within our "Movies" component by choosing their interests among sub-topics provided by our system. In order to compare the results before and after adding engagement to our components, we decided to use the average number of conversation turns as a metric to measure the interaction quality because conversation turns are the main reflection of users' interest, whether to continue on with the conversation or not. The comparison is conducted between before and after 7/8 for the Movies component, 6/23 for both the Food and News component, of which the dates are the time we added engagement respectively. The comparison results are shown in Figure \ref{fig:Engagement}.

The average number of conversation turns for the Movies component has increased by about 71.0\% (from 1.554 to 2.667) after adding the engagement technique, 45.0\% (from 1.280 to 1.850) for the Food component, and 50.0\% (from 1.830 to 2.745) for the News component. As a result, it is very clear that enabling in-depth conversation directly improve on overall conversational length, thus reflecting that users are willing to continue with the conversation than before.

Two-tailed T-test for two samples with unequal variance was used. For the Movie component, the p-value is <0.001, which supports the result of significant improvement. The same tests were conducted for Food and News components, with p-values both <0.001, indicating significant differences. This analysis confirms our conclusion that the engagement mechanism is crucial to the overall system performance.

\commentout{\todo{Figures 9 and 10 can be combined into single graph to save space, with conversation turns on secondary Y axis}}

\subsubsection{Proactivity}

Whenever an ODCS system conducts a conversation with users, handling situations when users are unsure of the next topic is significant in terms of maintaining the natural conversational flow. In other words, proactive behavior is a reflection of how intelligent an ODCS system is, since smart suggestions can directly influence the interaction quality by helping users to focus on and continue with the overall conversation.


    \line(541,0){240}
    \begin{dialogue}
        \speak{Ericson} How are you doing today?
        \speak{User} I'm doing great. Let's talk about Mars
        \speak{Ericson} We would normally refer to a planets diameter in relation to earth. Mars as a diameter 0.53 that of earth. Therefore mars has an equatorial diameter of 6760KM. I have something interesting about Cosmos, would you like to hear about it?
        \speak{User} Sure, tell me about that.
        \speak{Ericson} Here is the latest headline for science fans about outer space. Jim Bridenstine to Be Nominated by Trump to Lead NASA. The former Navy pilot is in his third term in Congress and would be the first elected official to serve as NASA administrator if confirmed by the Senate. Would you like another one?
        \speak{User} Who is Jim Bridenstine?
        \speak{Ericson} Here is what I got from Wikipedia. James Frederick Bridenstine is an American politician who has been the United States Representative for Oklahoma's 1st congressional district, based in Tulsa, since 2013. I have something interesting about celebrity, would you like to hear about it?
        \speak{User} I love celebrity news. Tell me something about it
        \speak{Ericson} Kim Kardashian West slammed for Jackie Onassis photoshoot. Would you like another one?
    \end{dialogue}
\line(552,0){240}

\begin{figure}[h]
\centering
  \includegraphics[width=1.0\columnwidth]{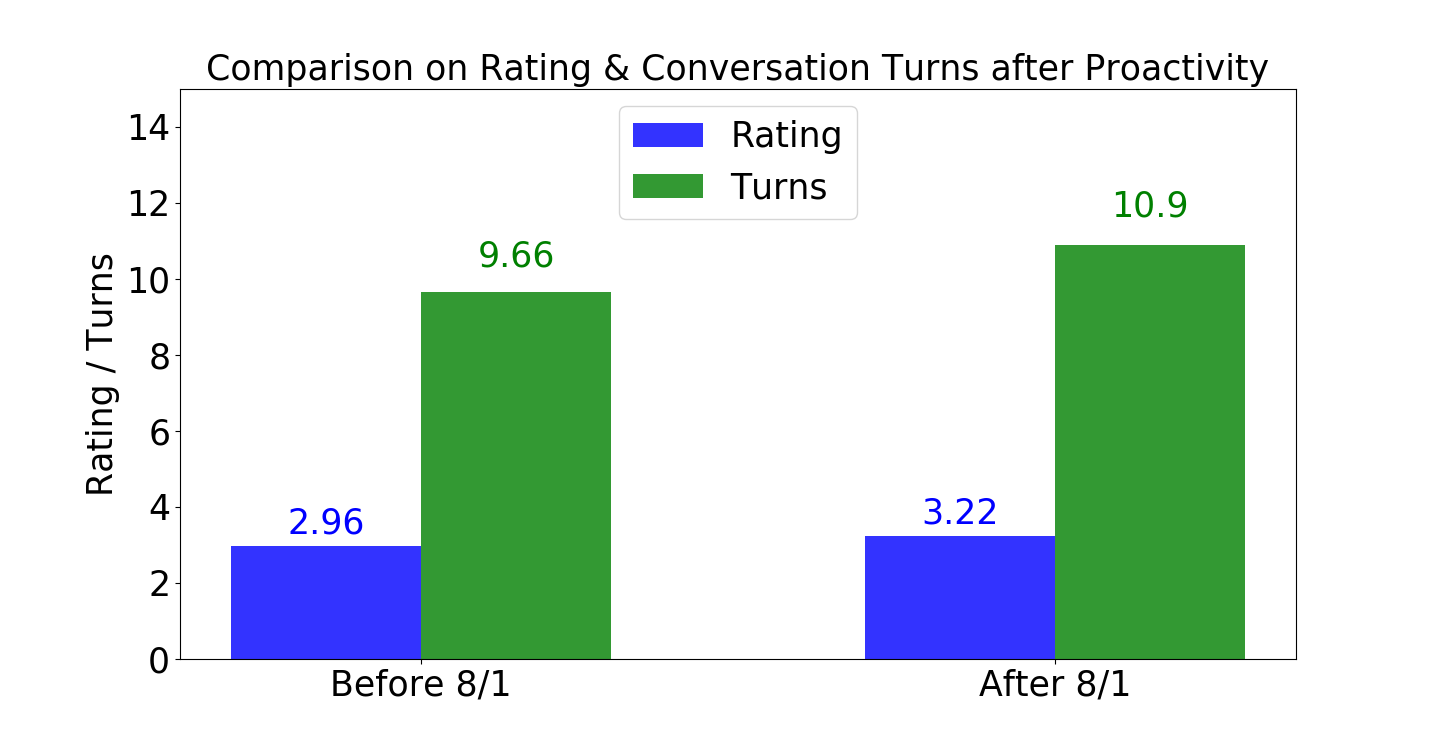}
  \caption{Average user rating and turns before and after proactivity completed}
  \label{fig:Proactive_Combined}
\end{figure}

Our system utilizes the entity classifier to detect any possible interest based on detected entities, and maps it to our available topics for suggestions, which successfully improved users' satisfaction. We used two metrics to measure the improvement, which are user ratings and average number of conversation turns before and after 8/1, the date we added the smart topic suggestion and proactive behavior to our system. The results are shown in Figure \ref{fig:Proactive_Combined}. 

As shown in the figures, improvement on the average number of conversation turns is 12.8\% (from 9.66 to 10.9) and improvement on user ratings is 8.6\% (from 2.962 to 3.218). While improvements from engagement focused on changes in component-wise level, we focused on overall changes for proactivity because proactivity mainly helps DSM to smooth the conversational transition and general information flow. 



Two-tailed T-test for two samples with unequal variance was used to evaluate the contribution of proactivity utilizing both averaged rating data and the number of conversation turns before and after proactivity was added. The p-values for the differences were both <0.001, which confirms that the differences were significant and that the proactivity significantly improved performance. 

\subsection{Qualitative Analysis: User Free-form Feedback}

Users' free-form feedback provides their subjective opinions on the interaction experience. It provides more details conforming to ratings accordingly. Upon requests, only a subset of participants provided their feedback, but common trends emerge and suggest for future directions of improvement. Among all the comments, 62\% of the comments are positive, 19\% negative and 19\% neutral. We divided a total of 944 feedback into 4 major categories with examples, as shown below.

\begin{itemize}
\item General Attitude : General positive or negative feedback.
	\begin{itemize}
    \item "I thought everything was good."
    \item "It did not have any exciting features."
    \end{itemize}

\item Favorite Feature : Feedback on features users enjoyed.
	\begin{itemize}
    \item "The social bot was good at giving a list of its topics and at responding and asking questions and giving further information."
    \item "That was a good move to get me to answer a questionnaire and you were really really up to date on the breaking news. In general, very good at appropriate responses to the human and being quite interesting, thank you."
    \item "This social bot was very very knowledgeable and intelligent and it actually responded to some of my questions..."
	\end{itemize}

\item Problem Identification : Feedback on specific problems.
	\begin{itemize}
    \item "Well she went straight into learning about facts and telling me about them which is really interesting but i think she should pick up on more small talk like how is your day what's your name stuff like that but overall it was very good good job."
    \item "I felt that the bot was a little too pushy on topics it wants me to talk about."
	\end{itemize}

\item Irrelevant Comments : Irrelevant feedback.
	\begin{itemize}
    \item "You need to let her listen more cause i want to do a quiz and it wouldn't let me."
    \item "Yes i think this bot is ummm."
    \end{itemize}
\end{itemize}

The general attitude category is composed of short positive or negative sentiments, providing us with a high-level overview of our system's performance. The favorite feature category summarizes features that are positively embraced by users, supporting us in confidence in components composing the current system. Problem identification category provides us with detailed information on unsatisfactory functions. Lastly, irrelevant comments are mostly random utterances that are irrelevant to the system performance. Among identified problems, we summarized the most frequently addressed topics as well, as shown in Table \ref{Feedback_table}.

\begin{table}[htbp]
\centering
\begin{tabular}{l}
  \toprule
  1. Proactive suggestions can be irrelevant and repetitive \\
  2. Conversation is not natural enough \\
  3. Intent detection problems \\
  4. Information are sometimes irrelevant \\
  5. The need of more entertaining and small-talk elements \\
  \bottomrule
\end{tabular}
\caption{Top 5 Commonly Addressed Topics.}
\label{Feedback_table}
\end{table}

The qualitative free-form feedback from the users confirms our quantitative results. Users frequently addressed comments about proactivity, intent classification, information relevance, topic coverage, and natural flow of conversation. We can deduce that these topics are the areas that are directly related to user experience and satisfaction of an ODCS system. These aspects provide promising directions on how to improve the overall human-computer interactions for future ODCS agents.

\section{Limitations}
The main limitation of the reported study was that it was performed in a real-time setting. Thus, some of the system component changes may have potential effects on other components. Unfortunately, due to the limited time frame available for the user study, some of the changes made to classifiers, engagement, and proactivity components were completed within a short time period, thus making it challenging to exclusively attribute ratings to individual component updates. Nevertheless, the effects of major improvements of the intent classifier, engagement, and proactive topic recommendation were significant enough to be credited for the improvements to the overall user experience.

\section{Conclusions and Future Work}
\label{sec:conclusions}

\commentout{\todo{Brief summary of the paper contributions and research questions you attempted to answer}
\todo{Next, explain how the paper answered the three research questions RQ1-3 in introduction. Be explicit -- say how your findings/results/system addressed each of the questions.}}

In this paper, we described an open-domain conversational search agent prototype that has been interacting with a sufficient number of real users. Three significant factors are studied on their effects on the interactions. 


\commentout{
\begin{itemize}
\item RQ1: what components and functions should a conversational search agent include?
\item RQ2: what factors or features impact crucially on the performance of a conversational search agent?
\end{itemize}
}

Based on our conversation analysis, our initial answer to RQ1 is that an open-domain conversational search system requires two essential functions, information retrieval/question answering, and also careful conversational transition modeling. These two functions meet users' needs for obtaining desired information while naturally support progressing through a conversation. These two essential functions are organically incorporated featuring three factors, intent classification, engagement, and proactivity.

Our analyses also show that, as an answer to RQ2, the most crucial part of an ODCS system is to have a very accurate user intent detection component and a comprehensive dialogue manager to control the conversation flow. Since our information retrieval components have different mechanisms as well as varying topic coverages and depths, interactions with users can be jeopardized if the classifier or dialogue management fails. We spent substantial efforts on improving our classification system, starting from a simple classifier and leading to a hierarchical classification system, as well as developing carefully designed logical flows for the suggested conversation transitions.
\commentout{\todo{``Engagement'' here is overloaded. Do you mean ``proactivity''? We use engagement to describe user's engagement with the system}}
We believe that modeling proactive behavior and enhancing the engagement will further increase the quality of interaction between the user and agent. 

\commentout{
\begin{itemize}
\item RQ3: How could the users benefit from conversational search compared to traditional Web search, and for which tasks?
\end{itemize}
}
To summarize our findings, we have shown that a conversational search system such as the \SYSTEM, presented in this paper, could allow users to retrieve information more effectively, and with higher satisfaction, compared to using traditional Web search and recommendations, especially in restricted communication settings such as voice or text. The main reason, we confirmed experimentally, is providing practivity and interaction, which allows the user to more naturally explore and retrieve content of interest through the dialogue with the system instead of putting the burden solely on the user to formulate and query the system. 

Our current and future work will focus on improving the conversational engagement and proactivity of \SYSTEM to enhance the conversation interactions, as well as adding more in-depth support for additional topical domains. 


\section{Acknowledgments}
We gratefully acknowledge the computational and technical support from the Amazon Alexa team during the Alexa Prize competition, which made the reported experiments possible.

\balance{}

\bibliographystyle{SIGCHI-Reference-Format}
\bibliography{sample}


\begin{thebibliography}{00}


\ifx \showCODEN    \undefined \def \showCODEN     #1{\unskip}     \fi
\ifx \showDOI      \undefined \def \showDOI       #1{{\tt DOI:}\penalty0{#1}\ }
  \fi
\ifx \showISBNx    \undefined \def \showISBNx     #1{\unskip}     \fi
\ifx \showISBNxiii \undefined \def \showISBNxiii  #1{\unskip}     \fi
\ifx \showISSN     \undefined \def \showISSN      #1{\unskip}     \fi
\ifx \showLCCN     \undefined \def \showLCCN      #1{\unskip}     \fi
\ifx \shownote     \undefined \def \shownote      #1{#1}          \fi
\ifx \showarticletitle \undefined \def \showarticletitle #1{#1}   \fi
\ifx \showURL      \undefined \def \showURL       #1{#1}          \fi

\bibitem{bohus2009ravenclaw}
{Dan Bohus} {and} {Alexander~I Rudnicky}. 2009.
\newblock \showarticletitle{The RavenClaw dialog management framework:
  Architecture and systems}.
\newblock {\em Computer Speech \& Language\/} {23}, 3 (2009), 332--361.
\newblock


\bibitem{bos2003dipper}
{Johan Bos}, {Ewan Klein}, {Oliver Lemon}, {and} {Tetsushi Oka}. 2003.
\newblock \showarticletitle{DIPPER: Description and Formalisation of an
  Information-State Update Dialogue System Architecture.}. In {\em SIGDIAL
  Workshop}. 115--124.
\newblock


\bibitem{friedman2001greedy}
{Jerome~H Friedman}. 2001.
\newblock \showarticletitle{Greedy function approximation: a gradient boosting
  machine}.
\newblock {\em Annals of statistics\/} (2001), 1189--1232.
\newblock


\bibitem{jafarpour2010filter}
{Sina Jafarpour}, {Chris Burges}, {and} {Alan Ritter}. 2010.
\newblock \showarticletitle{Filter, rank, and transfer the knowledge: Learning
  to chat}.
\newblock {\em Advances in Ranking\/}  {10} (2010).
\newblock


\bibitem{kim2014convolutional}
{Yoon Kim}. 2014.
\newblock \showarticletitle{Convolutional neural networks for sentence
  classification}.
\newblock {\em arXiv preprint arXiv:1408.5882\/} (2014).
\newblock


\bibitem{lemon2006isu}
{Oliver Lemon}, {Kallirroi Georgila}, {James Henderson}, {and} {Matthew
  Stuttle}. 2006.
\newblock \showarticletitle{An ISU dialogue system exhibiting reinforcement
  learning of dialogue policies: generic slot-filling in the TALK in-car
  system}. In {\em Proceedings of the Eleventh Conference of the European
  Chapter of the Association for Computational Linguistics: Posters \&
  Demonstrations}. Association for Computational Linguistics, 119--122.
\newblock


\bibitem{li2016learning}
{Jiwei Li}, {Alexander~H Miller}, {Sumit Chopra}, {Marc'Aurelio Ranzato}, {and}
  {Jason Weston}. 2016.
\newblock \showarticletitle{Learning Through Dialogue Interactions}.
\newblock {\em arXiv preprint arXiv:1612.04936\/} (2016).
\newblock


\bibitem{marietto2013artificial}
{Maria das Gra{\c{c}}as~Bruno Marietto}, {Rafael~Varago de Aguiar}, {Gislene
  de~Oliveira Barbosa}, {Wagner~Tanaka Botelho}, {Edson Pimentel}, {Robson
  dos~Santos Fran{\c{c}}a}, {and} {Vera~L{\'u}cia da Silva}. 2013.
\newblock \showarticletitle{Artificial intelligence markup language: A brief
  tutorial}.
\newblock {\em arXiv preprint arXiv:1307.3091\/} (2013).
\newblock


\bibitem{mikolov2013efficient}
{Tomas Mikolov}, {Kai Chen}, {Greg Corrado}, {and} {Jeffrey Dean}. 2013.
\newblock \showarticletitle{Efficient estimation of word representations in
  vector space}.
\newblock {\em arXiv preprint arXiv:1301.3781\/} (2013).
\newblock


\bibitem{nigam1999using}
{Kamal Nigam}, {John Lafferty}, {and} {Andrew McCallum}. 1999.
\newblock \showarticletitle{Using maximum entropy for text classification}. In
  {\em IJCAI-99 workshop on machine learning for information filtering},
  Vol.~1. 61--67.
\newblock


\bibitem{papaioannou2017combining}
{Ioannis Papaioannou} {and} {Oliver Lemon}. 2017.
\newblock \showarticletitle{Combining Chat and Task-Based Multimodal Dialogue
  for More Engaging HRI: A Scalable Method Using Reinforcement Learning}. In
  {\em Proceedings of the Companion of the 2017 ACM/IEEE International
  Conference on Human-Robot Interaction}. ACM, 365--366.
\newblock


\bibitem{serban2016building}
{Iulian~Vlad Serban}, {Alessandro Sordoni}, {Yoshua Bengio}, {Aaron~C
  Courville}, {and} {Joelle Pineau}. 2016.
\newblock \showarticletitle{Building End-To-End Dialogue Systems Using
  Generative Hierarchical Neural Network Models.}. In {\em AAAI}. 3776--3784.
\newblock


\bibitem{ultes2017pydial}
{Stefan Ultes}, {Lina M~Rojas Barahona}, {Pei-Hao Su}, {David Vandyke}, {Dongho
  Kim}, {Inigo Casanueva}, {Pawe{\l} Budzianowski}, {Nikola Mrk{\v{s}}i{\'c}},
  {Tsung-Hsien Wen}, {Milica Gasic}, {and} {others}. 2017.
\newblock \showarticletitle{Pydial: A multi-domain statistical dialogue system
  toolkit}.
\newblock {\em Proceedings of ACL 2017, System Demonstrations\/} (2017),
  73--78.
\newblock


\bibitem{wagner2015lambda}
{Tim Wagner}. 2015.
\newblock Running Arbitrary Executables in AWS Lambda.
\newblock
  \url{https://aws.amazon.com/blogs/compute/running-executables-in-aws-lambda/}.
    (2015).
\newblock


\bibitem{yao2015attention}
{Kaisheng Yao}, {Geoffrey Zweig}, {and} {Baolin Peng}. 2015.
\newblock \showarticletitle{Attention with intention for a neural network
  conversation model}.
\newblock {\em arXiv preprint arXiv:1510.08565\/} (2015).
\newblock


\end{thebibliography}

\end{document}